\documentclass{article}
\usepackage[preprint]{neurips_2026}
\usepackage[utf8]{inputenc} 
\usepackage[T1]{fontenc}    
\usepackage{hyperref}       
\usepackage{url}            
\usepackage{booktabs}       
\usepackage{amsfonts}       
\usepackage{nicefrac}       
\usepackage{microtype}      
\usepackage{xcolor}         
\usepackage{wrapfig}
\usepackage{algorithm}
\usepackage{enumitem}
\usepackage{algorithmic}
\usepackage{amsmath}
\usepackage{multirow}
\usepackage{microtype}
\usepackage{amssymb}
\usepackage{mathtools}
\usepackage{subcaption}
\usepackage{caption}
\usepackage{amsthm}
\usepackage{hyperref}
\definecolor{mycitecolor}{RGB}{0,90,180}
\usepackage[capitalize,noabbrev]{cleveref}
\hypersetup{
    colorlinks=true,
    citecolor=cyan,
    linkcolor=red,
    urlcolor=blue
}
\renewcommand{\algorithmiccomment}[1]{\hfill // #1}

\usepackage{amsmath,amsfonts,bm}

\def\eqref#1{equation~\ref{#1}}

\def\1{\bm{1}}

\def\pdg{{\texttt{\textup{PDG}}}}

\def\dllma{{Attn-Sampler}}
\def\mask{{\mathtt{MASK}}}

\def\vepsilon{{\boldsymbol{\epsilon}}}
\def\vtheta{{\mathbf{\theta}}}

\def\vb{{\mathbf{b}}}
\def\vc{{\mathbf{c}}}

\def\ve{{\mathbf{e}}}

\def\vh{{\mathbf{h}}}
\def\vu{{\mathbf{i}}}

\def\vk{{\mathbf{k}}}

\def\vp{{\mathbf{p}}}
\def\vq{{\mathbf{q}}}

\def\vs{{\mathbf{s}}}

\def\vu{{\mathbf{u}}}
\def\vv{{\mathbf{v}}}

\def\vx{{\mathbf{x}}}
\def\vy{{\mathbf{y}}}
\def\vz{{\mathbf{z}}}

\def\mA{{\mathbf{A}}}

\def\mW{{\mathbf{W}}}

\def\mZ{{\mathbf{Z}}}

\def\vtheta{{\bm{\theta}}}

\def\vb{{\bm{b}}}
\def\vc{{\bm{c}}}

\def\ve{{\bm{e}}}

\def\vh{{\bm{h}}}

\def\vk{{\bm{k}}}

\def\vp{{\bm{p}}}
\def\vq{{\bm{q}}}

\def\vs{{\bm{s}}}

\def\vu{{\bm{u}}}
\def\vv{{\bm{v}}}

\def\vx{{\bm{x}}}
\def\vy{{\bm{y}}}
\def\vz{{\bm{z}}}

\def\mA{{\bm{A}}}

\def\mW{{\bm{W}}}

\def\mZ{{\bm{Z}}}

\DeclareMathAlphabet{\mathsfit}{\encodingdefault}{\sfdefault}{m}{sl}
\SetMathAlphabet{\mathsfit}{bold}{\encodingdefault}{\sfdefault}{bx}{n}

\newcommand{\R}{\mathbb{R}}

\theoremstyle{plain}
\newtheorem{theorem}{Theorem}[section]
\newtheorem{proposition}{Proposition}[section]

\theoremstyle{plain}

\theoremstyle{plain} 

\theoremstyle{remark}

\title{Attention-Based Sampler for \\ Diffusion  Language Models}

\author{%
  Yuyan Zhou 
  \quad
  Kai Syun Hou 
  \quad
  Weiyu Chen\thanks{Corresponding author: \texttt{wchenbx@connect.ust.hk}} 
  \quad
  James Kwok \\
  Department of Computer Science and Engineering\\
  The Hong Kong University of Science and Technology \\
}

\begin{document}

\maketitle

\begin{abstract}
    Auto-regressive models (ARMs) have established a dominant paradigm in language modeling.
    However, their strictly sequential sampling paradigm imposes fundamental constraints on both inference efficiency and modeling flexibility.
    To address these limitations, diffusion-based large language models (dLLMs) have been proposed, offering the potential for parallel sampling and flexible language modeling.
    Despite these advantages, current dLLMs sampling strategies rely primarily on token level information, which fails to account for global sequence structure and often yields suboptimal results.
    In this paper, we study the sampling order selection problem from the perspective of log-likelihood maximization.
    We show that this problem is NP-hard and propose an optimal sampling-rank-based approximation that makes the objective computationally tractable.
    We further prove that the tractable objective is optimized by sampling tokens in descending order of their attention-matrix column sums.
    This finding provides a principled justification for attention-guided sampling and offers a theoretically grounded alternative to greedy search.
    We instantiate this theoretical insight in a new training-free sampling algorithm, termed \dllma, and further propose dynamic attention thresholding for practical acceleration.  
    Extensive experiments across multiple benchmarks validate the effectiveness of our proposed method, demonstrating that it achieves superior generation quality while enhancing the sampling parallelism.
\end{abstract}
\section{Introduction}
Auto-regressive models (ARMs)~\citep{achiam2023gpt, liu2024deepseek, dubey2024llama} provide a principled and widely adopted framework for language modeling. It factorizes the joint distribution of a sequence into a product of conditional probabilities under a fixed left-to-right order.
However, this approach requires strictly sequential, token-by-token sampling, imposing fundamental limitations on computational efficiency and modeling flexibility.

Recent works on diffusion large language models(dLLMs)~\citep{nie2025large, ye2025dream, zhu2025llada, khanna2025mercury, wu2025fast2} relax this constraint by allowing permutation-based factorizations of the joint distribution.
Instead of committing to a fixed left-to-right order, dLLMs define the sequence likelihood through flexible sampling permutations, enabling parallel prediction and order-agnostic generation. 
From a probabilistic perspective, this raises a central question:
\begin{center}
\textit{How to choose the tokens' sampling order so as to maximize 
    the likelihood of target sequence?}
\end{center}

Existing sampling strategies for dLLMs address this question through token-level greedy search, such as confidence-based~\citep{zhu2025llada}, margin-based~\citep{kim2025train} and entropy-based sampling algorithm~\citep{ye2025dream}. 
For parallel sampling, they typically rely on simple thresholding~\citep{wu2025fast,kim2025KLASS,ben2025accelerated} or top-$k$ selection~\citep{ye2025dream}.
While effective in practice, these greedy algorithms lack a clear theoretical justification and connection to the log-likelihood maximization of the target sequence.
\begin{wrapfigure}{r}{0.6\linewidth}
    \centering
    \vspace{-0.5em}
    \includegraphics[width=1\linewidth]{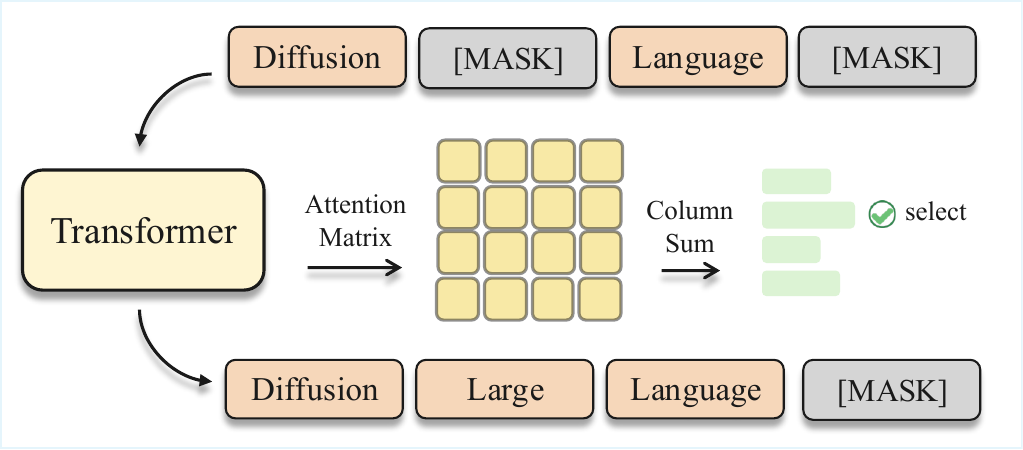}
    \caption{Overview of the \dllma\ algorithm. Our approach dynamically determines the sampling order for a masked sequence by leveraging the self-attention mechanism. We compute the column sums of the attention matrix as a proxy for token importance; tokens with higher cumulative attention scores are prioritized and decoded earlier in the sampling process.}
    \label{fig:teaser}
    \vspace{-1em}
\end{wrapfigure}
Consequently, these local selection samplers often yield suboptimal sampling trajectories and fail to maximize the log-likelihood of the target sequence.

In this work, we study the sampling order selection problem from a theoretical perspective. We formulate sampling order selection as an optimization problem that minimizes the gap between a practical permutation-based factorization likelihood and an ideal permutation-independent likelihood (where each token is predicted conditioned on all other tokens in the target sequence). Our main theoretical result shows that this gap is directly related to the token attention scores. However, such a direct optimization is computationally intractable. To alleviate this problem,
we propose an optimal 
approximation to make the objective computationally tractable and theoretically demonstrate that sampling tokens in descending order of their attention matrix column sums can minimize the tractable objective. This 
provides a formal bridge between the structural properties of self-attention and the log-likelihood maximization problem.

We instantiate this theoretical insight into a novel, training-free sampling algorithm called \dllma\ ({a}ttention-based sampling). 
Unlike existing methods that focus on the output probability space, \dllma\ utilizes the attention matrix of the transformer model to dynamically determine the most informative tokens to decode at each step. 
Additionally, we also propose dynamic attention thresholding, an adaptive mechanism that regulates sampling parallelism based on sequence-wide dependencies, leading to higher parallelism.

Our contributions can be summarized as follows:
\begin{enumerate}[leftmargin=*, align=left]
    \item 
    We show that the optimization problem of the sampling order selection is NP-Hard and propose an optimal 
    approximation to make the optimization computationally tractable. We prove that sampling in descending order of attention column sums minimizes the  tractable objective, thereby theoretically guaranteeing improved sampling quality.
    \item
    We propose \dllma, a training-free algorithm that leverages this theoretical insight. Through 
    an adaptive thresholding, it achieves efficient parallel sampling without compromising test accuracy.
    \item 
    We theoretically compare existing token-level samplers with our \dllma\ and identify the underlying causes of their differing performance in practice.
    \item 
    Extensive experiments demonstrate that \dllma\ consistently outperforms existing methods across multiple benchmarks in both accuracy and speed, establishing it as a robust new standard for dLLM inference.
\end{enumerate}

\section{Preliminary}
\subsection{Diffusion Large Language Models}

Given a sequence 
$\vx
= (x_0, x_1, \ldots, x_n) \in \R^n$,
where $x_0$ is the start (or context) token and each $x_i \in \mathcal{V}$ belongs to a finite vocabulary $\mathcal{V}$,
in auto-regressive models (ARMs), the probability of 
$\vx$
is factorized based on a fixed left-to-right ordering:
\begin{align*}
    p(\boldsymbol{x})
    = p(x_0)\prod_{i=1}^n p\big(x_i \mid x_0, \ldots, x_{i-1}\big).
\end{align*}
While computationally efficient, this fixed ordering imposes a rigid inductive bias that assumes the sequence structure is strictly unidirectional. However, in many settings, such as complex reasoning and code generation, the most informative tokens do not necessarily occur early in the sequence. Constraining the model to predict
$x_i$ solely based on its previous tokens will yield suboptimal learning and generation order, as it disregards dependency structures in the data. 

To address this limitation, 
discrete large language models 
(dLLMs)
\citep{sohl2015deep, meng2022concrete,austin2021structured, nie2025large} 
offer the potential for parallel sampling and flexible generation orders.

\subsubsection{Basics}
dLLMs
are a class of latent variable models with a forward noising process and a learned reverse denoising process.
Given data sample $\vx$ (also denoted
$\vx^0$ in 
the following, where the superscript denotes the time step),
the forward process progressively corrupts 
$\vx^0$
to a sequence of latent variables $\vx^1,\dots, \vx^T$ 
(with each $\vx^t\in \mathbb{R}^{n}$)
by iteratively applying noise. The
corresponding joint distribution is:
$
q(\vx^{1:T})=q(\vx^0)\prod^T_{t=1}q(\vx^t\mid \vx^{t-1}),
$
where $q(\vx^t \mid \vx^s) = \text{Cat}(\vx^t; \mathbf{Q}_t \vx^s)$ is the categorical distribution
and $\mathbf{Q}_t$
is the transition matrix at time step $t$.
On the other hand, the 
backward process learns to gradually denoise the masked sequence back to the original data
distribution by iteratively predicting masked tokens as $t$ moves from $T$ to $0$, with 
the corresponding joint distribution:
$
 p_{\vtheta}(\vx^{0:T}) \coloneqq p(\vx_{T}) \prod_{t=1}^{T} p_{\vtheta}(\vx^{t-1}|\vx^{t}).
$
The model parameter can be optimized by minimizing the negative log-likelihood of the clean data $\vx^0$.
When the model uses the absorbing kernel,
learning the denoising process leads to minimizing the Evidence Lower Bound (ELBO) of log likelihood~\citep{nie2025large}:
$\mathcal{L} (\vtheta) = -\mathbb{E}_{\vx^0, \vepsilon, t } \sum_{i=1}^N \mathbf{1}_{[x^t_i   = \mathtt{MASK}]} 
       \log p_{\boldsymbol{\vtheta}} (x^0_i \mid \vx^t),
$
where $x^0_i$ is the $i$th element of $\vx^0$.

\subsubsection{Sampling Process in dLLMs}
The sampling process involves iteratively unmasking tokens to reconstruct the original sequence. 
Early implementations use uniform sampling~\citep{austin2021structured}, where the token to be unmasked is selected at random. However, this ignores the model's varying levels of certainty across different positions.
To improve generation quality, research has shifted toward uncertainty-driven heuristics that utilize the model’s intrinsic predictive signals to guide the unmasking sequence~\citep{nie2025large, ye2025dream, kim2025train}. These strategies typically prioritize unmasking based on three metrics. (i) Confidence: Selecting tokens with the highest predicted probability for their most likely label~\citep{nie2025large}. (ii) Margin: Measuring the probability gap between the top two candidates 
\citep{kim2025train}. (iii) Entropy: Quantifying the overall predictive uncertainty across the entire distribution~\citep{ye2025dream}.
Recent advancements have further refined these heuristics by introducing adaptive samplers, which dynamically adjust the sampling process based on confidence~\citep{wu2025fast}, entropy~\citep{ben2025accelerated}, or KL-divergence~\citep{kim2025KLASS} to balance generation speed and sampling quality.

Each sampler specifies a generation order, represented by a permutation \(\pi\).
Existing dLLMs typically adopt block-wise 
\citep{arriola2025block} 
or semi-autoregressive sampling~\citep{nie2025large},
in which the permutation is defined at each block.
Let $\pi$ denote a permutation of the index set 
$\{1,\ldots,b\}$
within a block, such that
token
$i$ is decoded at step $\pi(i)$.
A permutation-based chain rule~\citep{kim2025train, chen2020generative} is used in dLLMs, which factorizes the joint distribution 
in a block with $b$ tokens
as:
\begin{align}
    p(\boldsymbol{x})
    = p(x_0)\prod_{i=1}^b p\big(x_{\pi(i)} \mid \boldsymbol{x}_{\pi(<i)}\big).
    \label{eq:permutation_likelihood}
\end{align}
Here, $\pi(<i)=\{j\in\{1,\ldots,b\}:\pi(j)<\pi(i)\}$ is the set of 
token indices
decoded before token $i$, and 
$\boldsymbol{x}_{\pi(<i)}$ is the corresponding set of tokens.
Thus, our goal is to find the permutation
which maximizes the log-likelihood:
\begin{align}
\label{eq:argmax_loglikelihood}
\pi^\ast =&\operatorname{\arg\max}_{ \pi }   \log  
\!\left(\! p_\vtheta({x}_0) \prod_{i=1}^b p_\vtheta({x}_{\pi(i)} 
| \boldsymbol{x}_{\pi(<i)})\!\right). 
\end{align}

\subsection{One-Layer Transformer with Softmax Attention}
\label{sec:xformer}

In the theoretical analysis of transformers,
the one-layer transformer
with softmax attention
has been commonly used
\citep{huang2024context, li2024one,siyu2024training, vaswani2017attention}.
Mathematically,
given a sequence 
$\vx
= (x_0, x_1, \ldots, x_n)$
of discrete random variables drawn from
distribution
$p_{\text{data}}$,
each token \( x_i \) is first mapped to a $d$-dimensional embedding $\ve_i = \text{Embed}(x_i) \in \mathbb{R}^d$. A positional encoding $\vp_i\in \mathbb{R}^d$ is added to form the input representation $\vh_i= \ve_i + \vp_i$.
Given 
$\{\vh_i\}_{i=0}^n$, a single-head self-attention layer 
has three $d\times d$ projection matrices: 
$\mW_{K},
\mW_{Q}$, 
and 
$\mW_{V}$, which
calculates
the query, key and value, respectively, as:
\begin{equation} \label{eq:attn}
\vq_i = \mW_Q \vh_i, \quad \vk_i = \mW_K \vh_i, \quad \vv_i = \mW_V \vh_i.
\end{equation}
The 
output representation $\vz_i$ is then computed as:
\begin{align}
\label{eq:attention}
\vz_i = \sum_{j} \mA_{ij} \vv_j, 
\end{align}
where 
$\{\mA_{ij} = \exp(\vq_i^T \vk_j / \sqrt{d} )/\sum_{j} \exp(\vq_i^T \vk_j / \sqrt{d})\}$
are the attention weights.
Finally, $\mZ = [\vz_1,\dots,\vz_n]$ is decoded to token probabilities by softmax, as:
\begin{equation}
 p_{\vtheta}(\vx) = \text{Softmax}(\mW\mZ + \vb\ve^\top),
 \label{output}
\end{equation}
where $\mW\in \mathbb{R}^{|\mathcal{V}|\times d}$, $\mZ\in \mathbb{R}^{d\times n}$, $\vb\in \mathbb{R}^{|\mathcal{V}|}$ and $\ve \in \mathbb{R}^n$ denotes the all-ones vector.
For notational simplicity, we use
$\vtheta$ to denote all the 
parameters
in this model.

\section{Proposed Method}
\label{sec:method}
In this section, we first formulate the determination of the sampling order in dLLMs as an optimization problem and show that it is NP-hard
(\cref{sec:order}). 
To make the optimization computationally tractable, we propose an optimal rank-only approximation and derive the optimality condition of the resulting tractable objective (\cref{sec:rk_only_approx}).
Based on this theoretical result, we propose \dllma,
which determines the sampling order based on the attention matrix (\cref{sec:algo}). 

\subsection{Sampling Order Selection as Optimization}
\label{sec:order}

Let $S=\{1,\ldots,b\}$ be the index set of $b$ tokens in a block.
On predicting each $x_i$, the model can only access $\left\{x_j \mid \pi(j)<\pi(i)\right\}$.
Here, we consider the 
ideal case where
the model can access all the remaining tokens
$\vx_{S\setminus i} = [x_1, \dots, x_{i-1}, x_{i+1}, \dots, x_b]$, with 
the 
corresponding likelihood
$            p_\vtheta(x_0)\prod_{i=1}^b
            p_\vtheta\big(x_{i} \mid \boldsymbol{x}_{S\setminus i}\big)$.
Obviously,
$\pi^\ast$ 
in (\ref{eq:argmax_loglikelihood})
is the same as
\begin{equation} \label{eq:pi-star}
\pi^\ast = \operatorname{\arg\min}_{ \pi }  \pdg_{\pi}(\vx, \vtheta),
\end{equation}
where
    \begin{align}
    &\pdg_{\pi}(\vx, \vtheta) =  \Bigg[
          \log \left(
            p_\vtheta(x_0)\prod_{i=1}^b
            p_\vtheta\big(x_{i} \mid \boldsymbol{x}_{S\setminus i}\big)
        \right)\notag 
         - \log \left(p_\vtheta({x}_0) \prod_{i=1}^b p_\vtheta({x}_{\pi(i)} \mid \boldsymbol{x}_{{\pi(<i)}})\right) \Bigg]
    \end{align}
will be called the  {\em permutation dependency gap}.
We interpret the first term 
on the RHS,
$\log \left(
p_\theta(x_0)\prod_{i=1}^b p_\theta\big(x_i \mid \boldsymbol{x}_{S\setminus i}\big)
\right)$, as the ``ideal'' (log) likelihood, where the model can access all non-target tokens when predicting each token.
Thus,
$\pi^\ast$ in (\ref{eq:pi-star})
finds the optimal permutation 
that minimizes the gap between this ``ideal'' likelihood 
and the permutation-based likelihood (the second term in RHS).

Let $\vv_m$ be the value obtained 
from (\ref{eq:attn})
on
a mask token,
and $\vv_j$ the value for token $x_j$ at position $j$.
Define $B = \sup_{j \in S} \Vert\vv_j - \vv_{m}\Vert_2$.
The following provides an upper bound for 
    $\pdg_{\pi}(\vx, \vtheta)$. Proof is in Appendix~\ref{appendix_a}.
    
\begin{proposition} \label{prop:pdg}
Under first-order approximation, 
$\pdg_{\pi}(\vx, \vtheta) 
    \leq    \sqrt{2}B   \Vert \mW\Vert_F \cdot
    \sum_{j=1}^b \sum_{i \in \pi(\leq j)} \mA_{ij} \coloneqq \mathrm{UB}(\vtheta, \vx, \pi)$,   
where
$\mA$
is the attention matrix 
in 
(\ref{eq:attention}), 
and $\Vert \mW\Vert_F$ is the Frobenius norm of $\mW$  in 
(\ref{output}).
\label{prop:upper_bound}
\end{proposition}

In Proposition~\ref{prop:pdg},
the term $\sum_{i \in \pi(\leq j)} \mA_{ij}$ 
can be
equivalently written as
$\sum_{i=1}^{b} \mathbf{1}({\pi(i)\leq\pi(j)}) \mA_{ij}$, where $\mathbf{1}(\cdot)$ is the indicator function.
This reformulation shows that 
$\mathrm{UB}(\vtheta, \vx, \pi)$ 
has the same structure as a {\em weighted linear ordering problem}, which is NP-hard~\citep{schiavinotto2004linear}.
Importantly, this expression also reveals that, for each token $j$, 
$\sum_{i\in \pi(\leq j)}\mathbf A_{ij}$ is not determined solely by its own {\em sampling rank}
$\pi(j)$,
but implicitly depends on the pairwise ordering structure
$\mathbf{1}(\pi(i)\leq\pi(j))$ across token pairs.
The computational hardness arises from this pairwise dependency: each cost term depends on the relative order of a token pair $(i,j)$, making the resulting optimization computationally intractable.

\subsection{Approximation for Tractable Optimization}
\label{sec:rk_only_approx}

To make the optimization problem tractable, we look for a function that
only uses 
each token $j$'s
sampling rank 
$\pi(j)$
to approximate the pairwise term $\sum_{i\in \pi(\leq j)}\mathbf A_{ij}$ in expectation, i.e.
\begin{align}
   \min_g
    \mathbb{E}_{\tilde{\pi} \mid \tilde{\pi}(j)=\pi(j)}
    \left(
    \sum_{i\in \tilde{\pi} (\leq j)} \mathbf A_{ij}
    -
    g(\pi(j))
    \right)^2,
    \label{eq:rk_solution}
\end{align}
where the expectation is over all permutations ($\tilde{\pi}$) subject to that token \(j\) has the given sampling rank $\pi(j)$.

\begin{proposition}[ ]
\label{thm:rank_only_approx}
For token $j$, the optimal $g$
in (\ref{eq:rk_solution}) is:
    $g^\ast(\pi(j))
    =
    \frac{\pi(j)-1}{b-1}
    \sum_{i=1}^{b}
    \mathbf 1(i\neq j)\mathbf A_{ij}$.
\end{proposition}
Proof is in Appendix~\ref{appendix_a2}.
Using the above
to replace
$\sum_{i \in \pi(\leq j)} \mA_{ij}$
in $\mathrm{UB}(\vtheta, \vx, \pi)$, we have
\begin{align*}
\widehat{\mathrm{UB}}(\vtheta,\vx,\pi) \equiv
\sqrt{2} B\Vert \mW\Vert_F\cdot \sum_{j = 1}^b \left(\frac{\pi(j)-1}{b-1} \sum_{i=1}^{b} \mathbf{1}(i\neq j) \mA_{ij}\right).
\end{align*}

Define the ``total attention" of token $j$ as $
s_j = \sum_{i=1}^b\mathbf{1}(i\neq j)  \mA_{ij}$.
The following Theorem shows that
$\widehat{\mathrm{UB}}(\vtheta,\vx,\pi)$
is minimized when tokens 
are decoded in descending total attention order. Proof is in Appendix~\ref{appendix_a3}.
\begin{theorem}
    \label{thm:best_order}
$\widehat{\mathrm{UB}}(\vtheta,\vx,\pi)$
is minimized when the 
permutation $\pi$ decodes 
tokens 
in descending 
total attention 
order.
\end{theorem}

\subsection{Algorithms}
\label{sec:algo}
\paragraph{Sequential Sampling.} 
\label{sec:single}

We now present \dllma\  
(\cref{alg:dllma}), a training-free sampling algorithm for dLLMs.
Given a prompt $\vc$ and block size
$b$, we first append $b$ mask tokens at the end of the prompt (step 3).
The model performs a forward pass 
to output the probability matrix $\vp\in \mathbb{R}^{b\times |\mathcal{V}|}$ and attention matrix $\mA$ with its diagonal elements removed (step 5).  
The total attention scores in $\vs \in \mathbb{R}^{b}$ are then computed
(step 6).
Based on Theorem~\ref{thm:best_order},
the masked tokens
are decoded in descending 
total attention
order
such that $s_{\pi(1)} \ge s_{\pi(2)} \ge \cdots \ge s_{\pi(b)}$ (step 7).
Finally, in steps 8-9, we identify and decode the token corresponding to the maximum total attention.

\paragraph{Parallel Sampling.}
\label{sec:parallel_algo}
Current parallel sampling methods \citep{ben2025accelerated, wu2025fast} employ static strategies on token level information,
such as fixed confidence/entropy thresholds, or top-$k$ token selection, to determine which tokens to decode simultaneously.
However, they neglect the global structure of the sequence,
leading to suboptimal generation performance as 
will be seen in the experiments (\cref{sec:acc}).

To address this issue, we extend the sequential sampling method in Section \ref{sec:single} to parallel sampling
(\cref{alg:dllma_parallel}).
Specifically, after computation of the total attention scores as in Algorithm~\ref{alg:dllma},
we 
use a probability threshold
$\tau$ to
partition the set of mask tokens into two subsets
(step 7):
(i) a candidate 
set, whose
tokens have probabilities  greater than or equal to $\tau$, and (ii) the non-candidate set with the remaining tokens. 
In the experiments, $\tau$ is set to 0.9.
Among tokens in the non-candidate set,
we identify the maximum total attention score 
as a ``dynamic threshold"
(step 8).
Only candidate tokens whose total attention scores 
exceed this threshold are decoded in the current time step (steps 9-11).   
This ensures that we only parallelize the sampling of the most important and independent tokens~\citep{wu2025fast}, thereby preserving generation quality. 

\begin{figure}[t]
\centering

\begin{minipage}[t]{0.48\linewidth}
\begin{algorithm}[H]
  \caption{\dllma\ (Sequential).}
  \label{alg:dllma}
  \begin{algorithmic}[1]
    \STATE {\bfseries Input:} Prompt $\mathbf{c}$, diffusion language model $f_{\vtheta}$, block size $b$.
    \REPEAT
    \STATE $\mathbf{c} \leftarrow \operatorname{Concat}(\mathbf{c}, b \times \mask)$
    \WHILE{$\mask \in \mathbf{c}$}
    \STATE $\vp, \mA \leftarrow f_{\vtheta}(\mathbf{c})$
    \STATE $s_i \leftarrow \sum_{j} \mA_{ji}$
    \STATE $u \leftarrow \arg\max_{i \in \text{masked}} s_i$
    \STATE $\hat{\mathbf{x}} \leftarrow \arg\max (\vp, \text{dim=1})$
    \STATE $\mathbf{c}_u \leftarrow \hat{\mathbf{x}}_u$
    \ENDWHILE
    \UNTIL{ $\langle\texttt{eos}\rangle \in \mathbf{c}$ }
  \end{algorithmic}
\end{algorithm}
\end{minipage}
\hfill
\begin{minipage}[t]{0.48\linewidth}
\begin{algorithm}[H]
  \caption{\dllma\ (Parallel).}
  \label{alg:dllma_parallel}
  \begin{algorithmic}[1]
    \STATE {\bfseries Input:} Prompt $\mathbf{c}$, diffusion language model $f_{\vtheta}$, block size $b$, threshold $\tau$.
    \REPEAT
    \STATE $\mathbf{c} \leftarrow \operatorname{Concat}(\mathbf{c}, b \times \mask)$
    \WHILE{$\mask \in \mathbf{c}$}
    \STATE $\vp, \mA \leftarrow f_{\vtheta}(\mathbf{c})$
    \STATE $s_i \leftarrow \sum_{j} \mA_{ji}$
    \STATE $\mathcal{I}_{\text{low}} \leftarrow \{i \mid \max p_i < \tau\}$
    \STATE $\gamma \leftarrow \max_{j \in \mathcal{I}_{\text{low}}} s_j$
    \STATE $\mathcal{U} \leftarrow \{i  \mid s_i > \gamma\}$
    \STATE $\hat{\mathbf{x}} \leftarrow \arg\max (\vp, \text{dim=1})$
    \STATE $\forall u \in \mathcal{U},\ \mathbf{c}_u \leftarrow \hat{\mathbf{x}}_u$
    \ENDWHILE
    \UNTIL{ $\langle\texttt{eos}\rangle \in \mathbf{c}$ }
  \end{algorithmic}
\end{algorithm}
\end{minipage}

\end{figure}
\section{Analysis of Different Samplers}

In this section, we analyze the relationships between traditional samplers and the proposed attention-based sampler.
In particular,
confidence-based sampler selects the token that has the highest predicted probability~\citep{nie2025large},
while entropy-based sampler
selects the token that minimizes predictive uncertainty~\citep{ye2025dream}.
Specifically,
let $\vp_i$ be the predictive probability distribution for token $i$, and $p_i = \max(\vp_i)$ the corresponding confidence (top probability). 
Confidence-based sampling selects $i^\ast = \arg\max
    (p_i)$,
while
entropy-based sampling
    selects
    token $i^\ast = \arg\min
    \mathcal{H}(\vp_i)$,
where $\mathcal{H}(\vp_i)$ is the predictive entropy.

First,
we show that
the proposed \dllma\ 
select the same token as confidence-based sampler under certain conditions.
First, we introduce a few symbols.
On predicting each $x_i$,
let
$\vz_i$ be the output 
from  (\ref{eq:attention})
when the model can only access $x_{\pi(<i)}$.
Analogous to
Section~\ref{sec:order}, we also consider the ideal case where
the model can access all remaining tokens $\vx_{S\setminus i}$, and let
the corresponding
output  from  (\ref{eq:attention})
be $\vz_i^\ast$.
Define the difference $\boldsymbol{\epsilon}_i = \vz_i - \vz_i^\ast$.

\begin{proposition} \label{prop:conf}
Confidence-based sampling selects the same token as \dllma\
under a first-order Taylor approximation,
when (i)
$\Vert \vepsilon_i\Vert_2  = h\left(\sum_{j \in \mathcal{M}} \mA_{ij}\right)$, where
$h$ is
a monotonically increasing function,
and $\mathcal{M}$ is the set of all mask tokens;
(ii)
$\sum_{j} \mA_{ji} = g\left(\sum_{j\notin \mathcal{M}} \mA_{ij}\right)$, where
$g$ is
a monotonically
increasing function; (iii)
the direction of $\vepsilon_i$ aligns with the negative gradient of confidence w.r.t. $\vz$; and (iv) In the ideal case, the predicted confidence is 1, and the gradient of the confidence score of token i with respect to $\vz_i^\ast$ becomes approximately constant.
    \label{thm:conf_equivalent}
\end{proposition}

Proof is in Appendix~\ref{appendix_a4}.
Next, the following Proposition shows that 
entropy-based sampling select the same token as 
confidence-based sampling.
\begin{proposition} \label{prop:entropy}
Entropy-based sampling selects
the same token 
$i^\ast$ 
as that selected by confidence-based sampling
    if either 
(i) 
$p_{i^\ast} = 1$, and $\forall i \neq i^\ast, p_i < 1$, or 
(ii) 
 $p_{i^\ast}>p_i$ for all $i\neq i^\ast$, and$
    -\mathcal H(\vp_{i^\ast})
    >
    \log p_i,
    \; \forall i\neq i^\ast,$ where $\mathcal H(\vp_{i^\ast})$ denotes the entropy of the predictive distribution at position $i^\ast$.
    \label{thm:conf_equivalent_entropy}
\end{proposition}
Proof is in Appendix~\ref{appendix_a5}.
In practice, 
the assumptions in Propositions~\ref{prop:conf} and
\ref{prop:entropy} 
rarely hold. Consequently, as will be empirically demonstrated in Section~\ref{sec:experiments}, confidence-based and entropy-based sampling often yield inferior results.

\begin{table*}[t]
\centering
    \caption{Testing accuracies of various sampling algorithms on different language models and evaluation tasks. 
    The best results are in bold, and the second-best are underlined.}
    \label{tab:accuracy}
    \vskip 0.15in
    \setlength{\tabcolsep}{4.0mm}{
\resizebox{1\textwidth}{!}{%
\begin{tabular}{clcccccc}
    \toprule
    & \textbf{Sampling method} & \textbf{GSM8K} & \textbf{Math} & \textbf{Humaneval} & \textbf{MBPP} 
    & \textbf{average} \\
    \midrule
    \multirow{7}{*}{Fast-dLLM v2 7B} 
    & KLASS & {83.24} & 50.96 & 54.88 & 30.69 & 54.94 \\   
    & EB-Sampler & 58.15 & 25.18 & 22.56 & 15.08 & 30.24 \\   
    & Fast-dLLM & 82.11 & 49.22 &  54.27 & 30.69 & 54.07 \\   
    & Margin Sampler & {82.94} & 49.92 & 51.22 & 30.42 & 53.63 \\
    & Entropy Sampler & 82.87 & \underline{51.92} & {55.49} & \underline{35.98}& {56.57} \\
    & Confidence Sampler & 82.71 & 50.96 & 54.27 & 30.69 & 54.66 \\
    & Attn-Sampler (Parallel) & \textbf{84.23} & 51.88 & \textbf{58.54} & \underline{35.98} & \underline{57.66} \\
    & Attn-Sampler (Sequential) & \underline{84.00} & \textbf{52.50} & \underline{57.93} & \textbf{36.24} & \textbf{57.67} \\
    \midrule
    \multirow{7}{*}{LLaDA-1.5 8B} 
    & KLASS & 73.69 & 38.76 & 40.85 & 50.79 & {51.02} \\   
    & EB-Sampler & 49.51 & 26.40 & {42.07} & 53.18 & 42.79 \\   
    & Fast-dLLM & \underline{74.91} & {39.62} & 40.24 & 48.68 & 50.86 \\  
    & Margin Sampler & 72.86 & 38.82 &  {41.46}  & 49.47 & 50.65 \\   
    & Entropy Sampler & 73.39 & 39.50 &  38.42  & 49.47 & 50.20 \\  
    & Confidence Sampler & 74.75 & \underline{39.64} & 40.85 & 48.68 & 50.98  \\  
    & Attn-Sampler (Parallel) & \underline{74.91} & 39.62 & \underline{42.68} & \textbf{53.97} & \underline{52.80} \\    
    & Attn-Sampler (Sequential) & \textbf{74.98} & \textbf{39.66} & \textbf{43.29} & \underline{53.44} & \textbf{52.84}\\  
    \midrule
    \multirow{7}{*}{Fast-dLLM v2 1.5B} 
    & KLASS & 61.64 & 32.22 & {37.20} & 27.78  & 39.71 \\  
    & EB-Sampler&  36.85 & 13.44  &  21.95 & 14.81  & 21.76\\  
    & Margin Sampler&  61.49 & 31.52  &  37.19 & 25.13  & 38.83 \\  
    & Entropy Sampler &  {62.55}  & 32.02 &  37.19 & 30.16 & {40.48} \\  
    & Confidence Sampler &  61.56 & \underline{32.24} & 37.19 & 27.78& 39.69 \\  
    & Attn-Sampler (Parallel) & \underline{62.62} & 31.02 & \textbf{39.63} & \textbf{30.96} & \underline{41.06} \\
    & Attn-Sampler (Sequential) & \textbf{62.70} & \textbf{32.56} & \textbf{39.63} & \underline{30.16} & \textbf{41.26}\\  
    \bottomrule
\end{tabular}
        }
}
\end{table*}
\section{Experiments}
\label{sec:experiments}
In this section, we validate the effectiveness of the proposed \dllma. We first compare its predictive accuracy and computational throughput against existing sampling methods, demonstrating that \dllma achieves state-of-the-art accuracy and a superior throughput-accuracy trade-off. Given that current transformers utilize multi-head attention across multiple layers, we further perform ablation studies on these components to demonstrate that aggregating information across all heads and layers maximizes performance.

\subsection{Setup}

\paragraph{Models and Hardware.} Evaluation is performed on two representative dLLMs: Fast-dLLM v2 (with 1.5B and 7B parameters) \citep{wu2025fast2} and LLaDA-1.5 8B \citep{zhu2025llada}. 
All experiments are run on a single NVIDIA A6000 GPU.

\paragraph{Benchmark Datasets.} 
We utilize two categories of benchmarks:
(i) mathematical reasoning: GSM8K \citep{cobbe2021gsm8k} and MATH \citep{2019arXiv};
(ii) code generation: HumanEval \citep{chen2021evaluating} and MBPP \citep{austin2021program}.

\begin{figure}[thb]
    \centering
    \begin{minipage}[t]{0.48\linewidth}
        \centering
        \includegraphics[width=\linewidth]{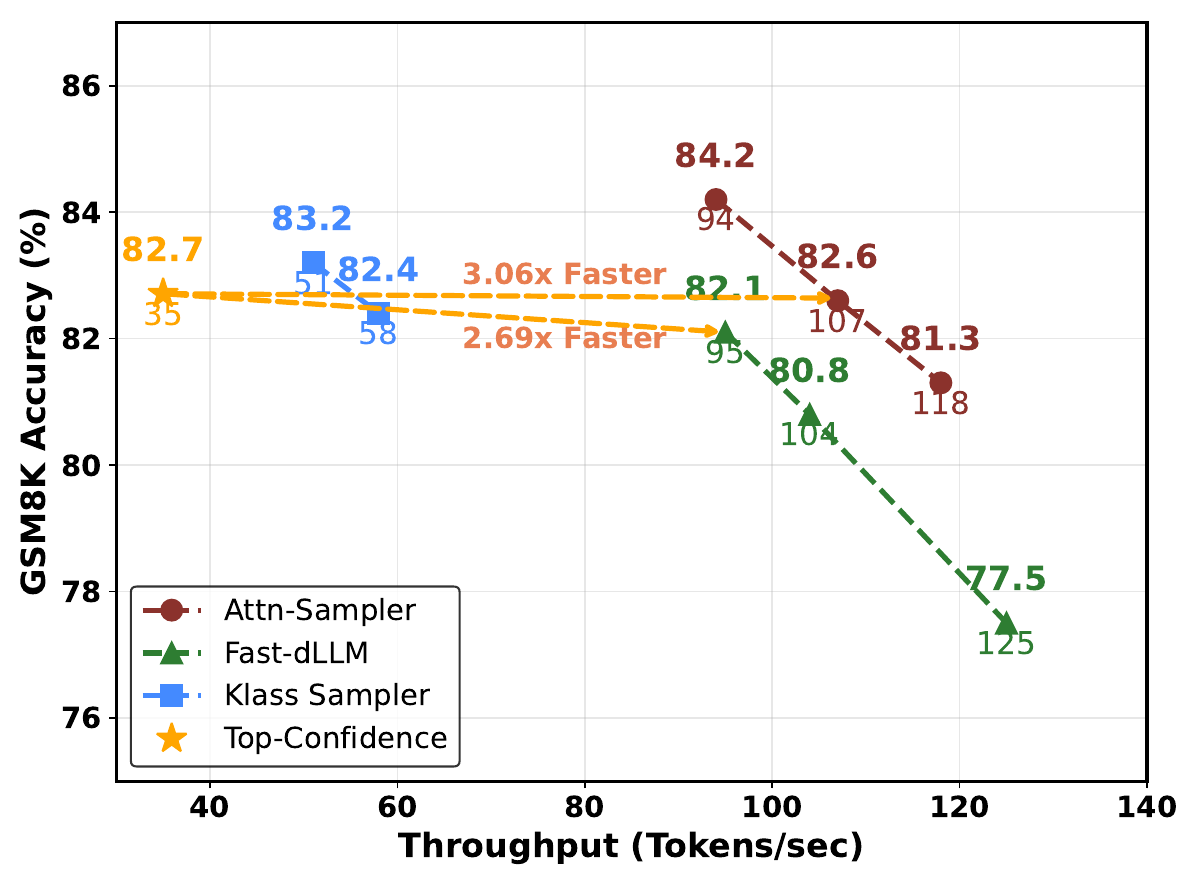}
        \captionof{figure}{Comparison of throughput vs. accuracy across various parallel sampling methods. Our \dllma\ defines a superior Pareto front compared to existing state-of-the-art parallel samplers.}
        \label{fig:accuracy_tps_comparison}
    \end{minipage}
    \hfill
    \begin{minipage}[t]{0.48\linewidth}
        \centering
        \includegraphics[width=\linewidth]{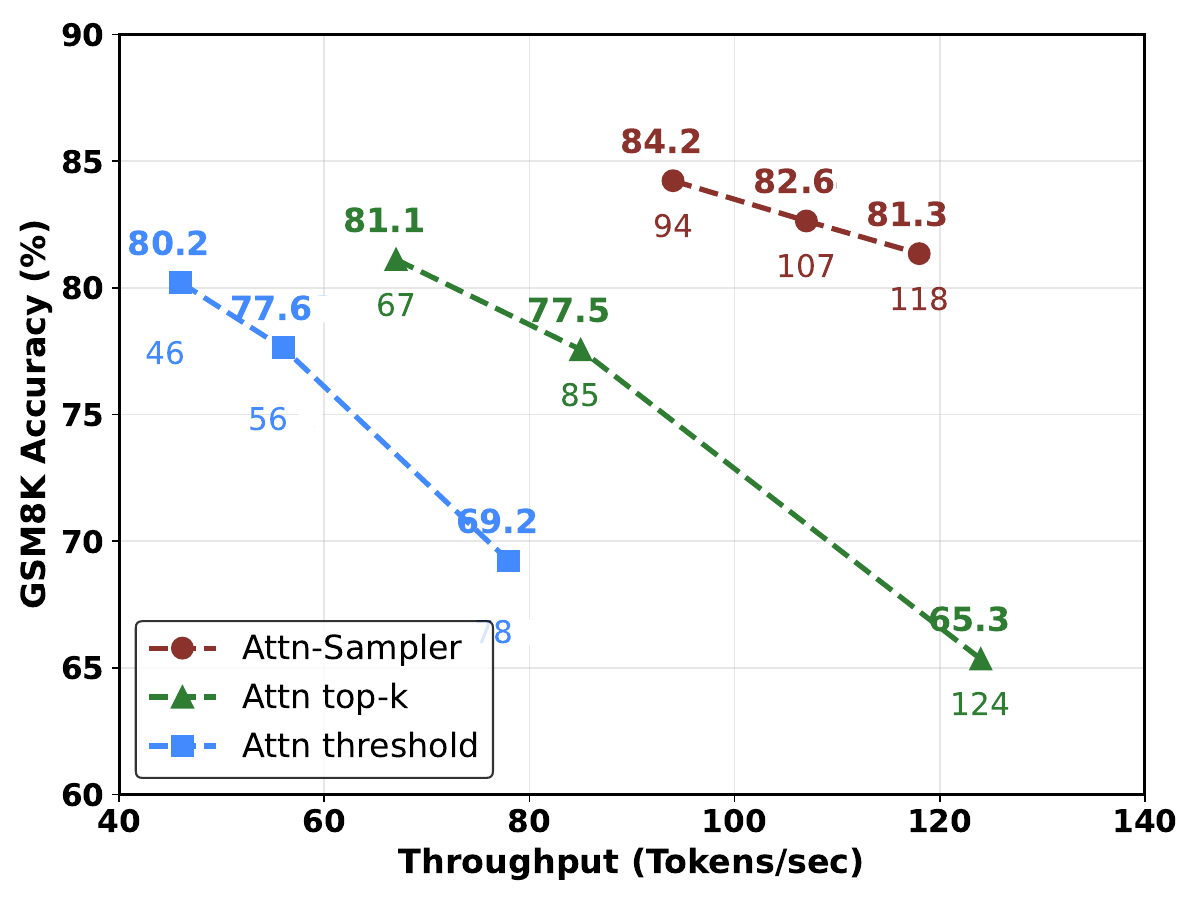}
        \captionof{figure}{Ablation study of attention sampling strategies. Our adaptive approach outperforms both top-$k$ and static thresholding, achieving a more efficient trade-off between speed and performance.}
        \label{fig:attn_ablation}
    \end{minipage}
\end{figure}

\subsection{Test Accuracy}
\label{sec:acc}
We compare the performance of \dllma\ against a suite of competitive sampling  baselines for dLLMs:
(i) top-1 confidence~\citep{nie2025large}, (ii) top-1 margin~\citep{kim2025train}, (iii) top-1 negative entropy~\citep{ye2025dream}, (iv) Fast-dLLM~\citep{wu2025fast}, (v) EB-Sampler~\citep{ben2025accelerated} and (vi) KLASS Sampler~\citep{kim2025KLASS}.

Table \ref{tab:accuracy} shows the accuracies of various samplers.
As can be seen, both sequential 
sampling 
and  parallel sampling 
of the proposed \dllma\ 
consistently achieve state-of-the-art results across all the 
model architectures and benchmarks
evaluated.
With the Fast-dLLM v2 7B, \dllma\ sequential sampling outperforms the top confidence sampler by 3.01\% on the average metric. Compared with the strongest baseline (entropy sampler), \dllma\ outperforms by 1.1\% on average, with a notable +2.44\% accuracy gain on 
HumanEval.
\dllma\
parallel sampling 
maintains accuracy levels comparable to those of sequential sampling, and even outperforming sequential sampling on GSM8K and Humaneval datasets.
With the different LLaDA-1.5 8B model, \dllma\  maintains the highest average accuracy (sequential:
52.84\%;
parallel: 52.80\%), demonstrating robustness across different diffusion scales.

The advantages of \dllma\ remain evident at the 1.5B parameter scale. With the Fast-dLLM v2 1.5B, \dllma\ achieves a 41.26\% average score, surpassing the KLASS and Fast-dLLM baselines by over 1.5 percentage points. This suggests that \dllma\ effectively captures the underlying distribution of the diffusion process, even when the base model capacity is limited.

\subsection{Inference Speed Comparison}

Using the Fast-dLLM v2 model on GSM8K,
we evaluate the trade-off between generation throughput 
(measured in tokens per second (TPS))
and task accuracy of various parallel sampling algorithms
(\dllma, 
Fast-dLLM, and
KLASS).
For \dllma, we vary the threshold $\tau$ to $0.9, 0.8$ and $0.7$;
for Fast-dLLM, we vary the confidence threshold to $0.9, 0.8$ and $0.7$. 
for KLASS, we vary the KL-divergence thresholds to $0.001$ and $0.01$.

\cref{fig:accuracy_tps_comparison}
shows the comparison of throughput-accuracy curves.
For comparison, we also show the result of
top-confidence sampling.
As can be seen, \dllma\ consistently shows a better Pareto front compared to the baselines.
While top-confidence sampling achieves an accuracy of $82.7\%$ with a throughput of $35$ TPS, the Fast-dLLM sampler provides a $2.69\times$ speedup ($95$ TPS) but at a slight drop in accuracy ($82.1$).
In contrast, at the same throughput of $95$ TPS, \dllma\ achieves a significantly higher accuracy of $84.2\%$.
Furthermore, \dllma\ can be configured for even higher efficiency, reaching $107$ TPS ($3.06\times$ acceleration) while maintaining $82.6\%$ accuracy, effectively matching the precision of the confidence-based baseline at triple the speed.
While the KLASS sampler reaches a competitive accuracy of $83.2\%$, its throughput is limited to $51$ TPS, making it substantially less efficient than our proposed method.
These results demonstrate that \dllma\ effectively bridges the gap between high-quality reasoning and high-throughput inference.

\subsection{Ablation Studies}
\paragraph{Dynamic Attention Thresholding}
\label{sec:attention_thresholding_ablation}
In this section, we ablate and validate the effectiveness of our proposed parallel attention sampler.
Static thresholding~\citep{wu2025fast} and top-$k$~\citep{ye2025dream} parallelization strategies can improve the throughput of confidence and entropy sampler.
However, we show that these strategies are insufficient for the requirements of \dllma.
To evaluate the effectiveness of our proposed dynamic attention thresholding mechanism, we conduct an ablation study using the Fast-dLLM v2 model on the GSM8K dataset.
We compare our approach against the two standard strategies: (i) Top-$k$ selection: Decode a fixed number of tokens corresponding to the highest attention scores, where $k$ is set to $2,3,4$. (ii) Static thresholding: Masks all attention weights falling below a predefined constant attention threshold, with threshold set as $1.0, 0.9, 0.8$.

\cref{fig:attn_ablation} shows the comparison of different attention based parallel sampling strategies.
As can be seen, \dllma\ exhibits a significantly more robust Pareto front than either baseline. 
While both the top-$k$ and static thresholding variants can increase throughput, they suffer from a sharp decline in accuracy as parallelism becomes higher. For example, increasing the throughput of the top-$k$ baseline from 67 TPS to 124 TPS results in an accuracy drop of approximately 15.77\% (from $81.12$ to $65.35$).
In contrast, \dllma\ maintains a high reasoning performance even at elevated speeds. At 118 TPS, our method still achieves $81.35\%$ accuracy, outperforming the top-$k$ method's accuracy at nearly half the speed (67 TPS).
The results suggest that static or fixed-count pruning fails to account for the varying importance of tokens across different diffusion steps. By adaptively scaling the threshold, \dllma\ effectively preserves the critical semantic information required for complex tasks like GSM8K while successfully eliminating redundant computations.

\begin{figure}[t]
    \centering
    \begin{minipage}[t]{0.48\linewidth}
        \centering
        \includegraphics[width=\linewidth]{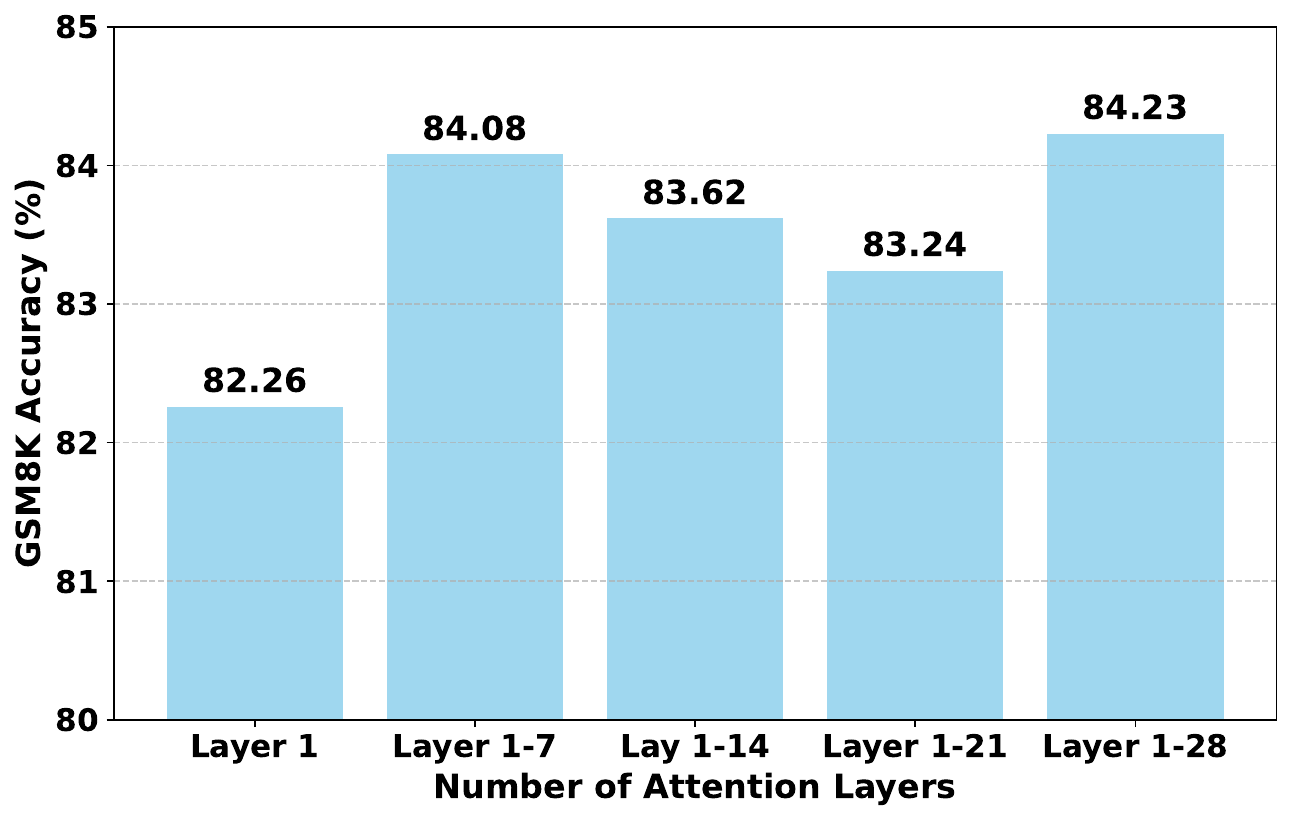}
        \captionof{figure}{Ablation study of test accuracy using different attention layers.}
        \label{fig:layer_ablation}
    \end{minipage}
    \hfill
    \begin{minipage}[t]{0.48\linewidth}
        \centering
        \includegraphics[width=\linewidth]{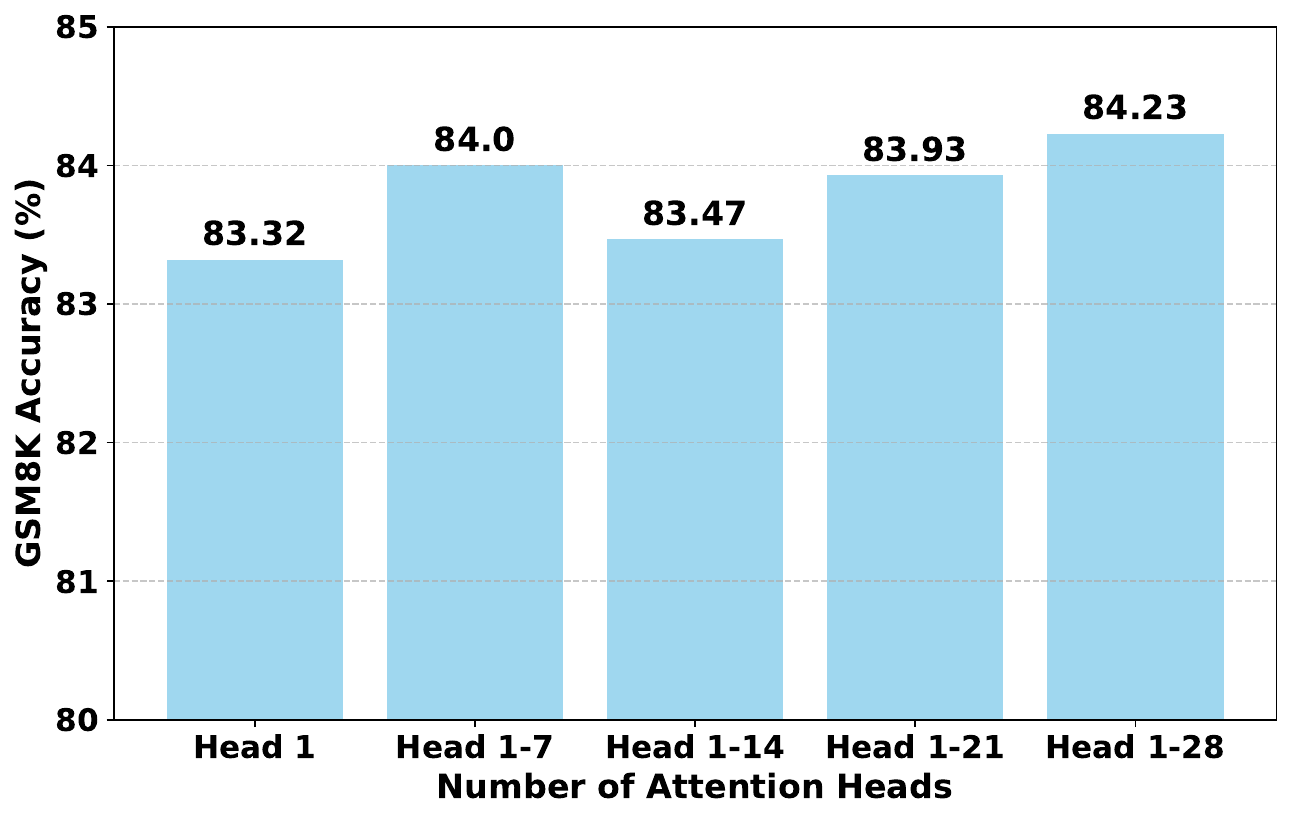}
        \captionof{figure}{Ablation study of test accuracy using different attention heads.}
        \label{fig:head_ablation}
    \end{minipage}

\end{figure}

\paragraph{Attention Layers and Heads}
To evaluate the contribution of the internal transformer components
to the sampling process, we conduct ablation studies by cumulatively increasing the number of active attention layers and heads.
Attention Layers: We first compare the performance of utilizing only the initial layers against different layer subsets 
up to the full model depth. 
\cref{fig:layer_ablation} shows the test accuracies using obtained.
As can be seen, relying solely on the first attention layer yields the lowest performance (82.26\%), whereas performance improves significantly when extending the scope to the first 7 layers (84.08\%) and attain highest when using the mean of all the layers.
Attention Heads: A similar cumulative study is conducted for attention heads within each layer. \cref{fig:head_ablation} shows the ablation study of test accuracy using different attention heads.
As can be seen, using only the first
head results in a baseline of 83.32\%, with performance scaling to 84.0\% as the first 7 heads are included.
For both layers and heads, the model achieves peak performance (84.23\% on GSM8K) only when the full architectural width and depth, specifically the mean of all 28 layers and heads are utilized. These results suggest that while early components capture fundamental features, the high-level semantic information and specialized representations distributed across the entire network are critical for maximizing model performance.

\section{Conclusion}
In this work, we introduced a principled approach to sampling order selection for dLLMs by framing it as a log-likelihood maximization problem.
This problem can be formulated as minimizing a ``permutation dependency gap''.
We show that this optimization problem is NP-hard, and propose an optimal sampling-rank-based approximation that makes the objective computationally tractable. We further prove that the resulting tractable objective is optimized by sampling tokens in descending order of their attention-matrix column sums.
Based on these findings, we propose \dllma, a training-free algorithm that uses attention scores to guide sequence generation. By incorporating 
dynamic thresholding, \dllma\ achieves superior generation quality and enhanced parallelism. We further theoretically compare existing token-level samplers with our Attn-Sampler and identify the underlying causes of their differing performance in practice.
Our results demonstrate that \dllma\ consistently outperforms existing methods, establishing a new, theoretically grounded standard for efficient dLLM sampling.

\bibliography{example_paper}
\bibliographystyle{plainnat} 

\appendix
\onecolumn
\section{Proofs}
\subsection{Proof of \cref{prop:upper_bound}}
\label{appendix_a}
First, rewrite the optimal permutation $\pi^*$ in (\ref{eq:pi-star}) as:
\[ \pi^*
 =     \arg\min_{\pi} \sum_{i=1}^b \left[ -\log p_\vtheta({x}_{\pi(i)} \mid \vx_{\pi(<i)}) + \log p_\vtheta(x_{\pi(i)} \mid \vx_{S \setminus \pi(i)}) \right]. \]
For the token at position $\pi(i)$,
let $\vz_{\pi(i)}$ be the representation 
obtained in (\ref{eq:attention}),
and
$\vz^\ast_{\pi(i)}$ be the representation in the
ideal case where the
model can access all remaining tokens.
From (\ref{output}),
$$\log p_\vtheta({x}_{\pi(i)} \mid \vx_{\pi(<i)}) = \log \frac{\exp (\mW\vz_{\pi(i)} + \vb)_{x_{\pi(i)}}}{\sum_k \exp (\mW\vz_{\pi(i)} + \vb)_k}\coloneqq\log p_\vtheta({x}_{\pi(i)} \mid \vz_{\pi(i)} ).$$
Let 
$\vu_{\pi(i)}
\equiv \mW\vz_{\pi(i)} + \vb$.
The gradient of $\log p_\vtheta({x}_{\pi(i)} \mid \vz_{\pi(i)})$
w.r.t. 
$\vu_{\pi(i)}$
is:
\[ \nabla_{\vu_{\pi(i)}} \log p_\vtheta({x}_{\pi(i)} \mid \vz_{\pi(i)} ) =  - \vp_{\pi(i)} + \vy_{x_{\pi(i)}}, \]
where $\vp_{\pi(i)}$ is the predictive probability vector 
for token $x_{\pi(i)}$
and  $\vy_{x_{\pi(i)}}\in\mathbb{R}^{|\mathcal{V}|}$ denotes the one-hot vector for token $x_{\pi(i)}$. Note that
\begin{align*}
    \left\Vert \nabla_{\vu_{\pi(i)}} \log p_\vtheta({x}_{\pi(i)} \mid\vz_{\pi(i)} ) \right\Vert_2^2
    = (1-p_{x_{\pi(i)}})^2 + \sum_{j\neq \pi(i)} p_{x_{j}}^2 \leq 2,
\end{align*}
i.e., its gradient norm is bounded and the Lipchitz constant of the log softmax function
is $\sqrt{2}$:
Given that the Lipchitz constant is $\sqrt{2}$ and $\vu_{\pi(i)}$ is linear in
$\vz_\pi(i)$ with weight $\mW$, the difference in log-probabilities is bounded by:
\begin{align*}
\log p_\vtheta({x}_{\pi(i)} \mid \vx_{S\setminus\pi(i)})- \log p_\vtheta({x}_{\pi(i)} \mid \vx_{\pi(<i)}) & = 
\log p_\vtheta({x}_{\pi(i)} \mid \vz_{\pi(i)}^\ast )- \log p_\vtheta({x}_{\pi(i)} \mid  \vz_{\pi(i)} )\\
&\leq \sqrt{2}  \|\mW\vz_{\pi(i)} -\mW\vz_{\pi(i)}^\ast \|_2 \\
&\leq \sqrt{2}\cdot \|\mW\|_F \cdot\|\vz_{\pi(i)} - \vz^\ast_{\pi(i)}\|_2.
\end{align*}

Next, we analyze the difference in the latent representations. In the Transformer architecture, the hidden state is a weighted sum of value projections $\vv_j$ mediated by attention weights $\mA_{ij}$. Let $\mathcal{M}$ be the set of indices for masked tokens. We can express $\vz_{\pi(i)}$ and $\vz^\ast_{\pi(i)}$
as:
$$\vz_{\pi(i)} = \sum_{j \notin \mathcal{M}} \mA_{\pi(i)j} \vv_j + \sum_{j \in \mathcal{M}} \mA_{\pi(i)j} \vv_m,$$
$$\vz_{\pi(i)}^\ast = \sum_{j \notin \mathcal{M}} \mA_{\pi(i)j} \vv_j + \sum_{j \in \mathcal{M}} \mA_{\pi(i)j} \vv_j.$$
By subtracting these expressions, the terms corresponding to non-masked tokens cancel out:$$\|\vz_{\pi(i)} - \vz^\ast_{\pi(i)}\|_2 = \left\| \sum_{j \in \mathcal{M}} \mA_{\pi(i)j} (\vv_m - \vv_j) \right\|_2.$$
Applying the Triangle Inequality and substituting the supremum of the value discrepancy $B = \sup_{j, m} \|\vv_j - \vv_m\|_2$, we obtain:
$$\|\vz_{\pi(i)} - \vz^\ast_{\pi(i)}\|_2 \leq \sum_{j \in \mathcal{M}} \mA_{\pi(i)j} \|\vv_m - \vv_j\|_2 \leq B \sum_{j \in \mathcal{M}} \mA_{\pi(i)j}.$$
In the context of the permutation $\pi$, the set of masked tokens $\mathcal{M}$ for the $i$-th step consists of all tokens that appear later
in the permutation, i.e., $\{j \mid \pi(j) \geq \pi(i)\}$. 
Summing over all $i$, we get the final upper bound:
\begin{eqnarray*}
    \text{PDG}_{\pi}(\vx)
    &\leq &\sqrt{2} \cdot \|\mW\|_F \cdot B \cdot \sum_{i=1}^b \sum_{j: \pi(j) \geq \pi(i)} \mA_{\pi(i)j}\\
    &= & \sqrt{2} \cdot \|\mW\|_F \cdot B \cdot \sum_{i=1}^b \sum_{j=1}^b \mathbf{1}(\pi(i)\leq \pi(j)) \mA_{ij}\\
    &= & \sqrt{2} \cdot \|\mW\|_F \cdot B \cdot \sum_{j=1}^b \sum_{i\in\pi(\leq j)}  \mA_{ij}.
\end{eqnarray*}

\subsection{Proof of \cref{thm:rank_only_approx}}
\label{appendix_a2}
For notational simplicity, denote $\sum_{i \in \hat{\pi} (\leq j)} \mA_{ij}$ as $Y_j(\tilde{\pi} )$.
For any function $g$, we have
\begin{eqnarray*}
\lefteqn{\mathbb E_{\tilde{\pi}  \mid \tilde{\pi}(j)=\pi(j)}
\left(
Y_j(\tilde{\pi} )-g(\pi(j))
\right)^2}\\
&= &
\mathbb E_{\tilde{\pi}  \mid \tilde{\pi}(j)=\pi(j)}
\left(
Y_j(\tilde{\pi} )
-
\mathbb E_{\tilde{\pi}  \mid \tilde{\pi}(j)=\pi(j)}[Y_j(\tilde{\pi} )]
+
\mathbb E_{\tilde{\pi}  \mid \tilde{\pi}(j)=\pi(j)}[Y_j(\tilde{\pi} )]
-
g(\pi(j))
\right)^2 \\
&= &
\mathbb E_{\tilde{\pi}  \mid \tilde{\pi}(j)=\pi(j)}
\left(
Y_j(\tilde{\pi} )-\mathbb E_{\tilde{\pi}  \mid \tilde{\pi}(j)=\pi(j)}[Y_j(\tilde{\pi} )]
\right)^2
+
\mathbb E_{\tilde{\pi}  \mid \tilde{\pi}(j)=\pi(j)}
\left(
\mathbb E_{\tilde{\pi}  \mid \tilde{\pi}(j)=\pi(j)}[Y_j(\tilde{\pi} )]-g(\pi(j))
\right)^2,
\end{eqnarray*}
where the cross term vanishes because
\[ \mathbb E_{\tilde{\pi}  \mid \tilde{\pi}(j)=\pi(j)}
\left[
Y_j(\tilde{\pi} )-\mathbb E_{\tilde{\pi}  \mid \tilde{\pi}(j)=\pi(j)}[Y_j(\tilde{\pi} )]
\right]
=0. \]
The first term is independent of $g$, while the second term is minimized when
\[    g(\pi(j))=\mathbb E_{\tilde{\pi}  \mid \tilde{\pi}(j)=\pi(j)}[Y_j(\hat{\pi} )]. \]
Therefore,
\[   g^\ast(\pi(j))=\mathbb E_{\tilde{\pi}  \mid \tilde{\pi}(j)=\pi(j)}[Y_j(\hat{\pi} )].\]

It remains to compute 
$\mathbb E_{\tilde{\pi}  \mid \tilde{\pi}(j)=\pi(j)}
\left[
Y_j(\tilde{\pi})
\right]$.
By definition,
\[ Y_j(\tilde{\pi} )
    =
    \sum_{i=1}^{b}
    \mathbf 1(i\in \tilde{\pi} (\leq j))
    \mathbf A_{ij}.  \]
Conditioned on $\tilde{\pi} (j) = \pi(j)$, each token $i (\neq j)$ is
included in $\pi(\leq j)$ with probability
\begin{align*}
    \Pr(i\in \tilde{\pi} (\leq j)\mid \tilde{\pi} (j)= {\pi} (j))
    =
    \frac{{\pi} (j) -1}{b-1} .
\end{align*}
Thus,
\begin{align*}
\mathbb E_{\tilde{\pi}  \mid \tilde{\pi}(j)=\pi(j)}
\left[
Y_j(\tilde{\pi})
\right]
&=
\sum_{i=1}^{b}
\mathbf 1(i\neq j)
\frac{\pi(j)-1}{b-1}
\mathbf A_{ij} \notag \\
&=
\frac{\pi(j)-1}{b-1}
\sum_{i=1}^{b}
\mathbf 1(i\neq j)
\mathbf A_{ij}.
\end{align*}

\subsection{Proof of \cref{thm:best_order}}
\label{appendix_a3}
First, 
recall that
\begin{equation} \label{eq:tmp}
\widehat{\mathrm{UB}}(\vtheta,\vx,\pi) 
= \sqrt{2}\Vert \mW\Vert_F\cdot \frac{B}{b-1}\cdot\sum_{j = 1}^b \left((\pi(j)-1)s_j\right),
\end{equation}  
where $s_j = \sum_{i=1}^b\mathbf{1}(i\neq j)  \mA_{ij}$ is the total attention of token $j$.
We first prove that when two 
tokens 
decoded at 
adjacent steps
are not in 
descending 
total attention 
order, interchanging their decoding orders can minimize $\widehat{\mathrm{UB}}(\vtheta,\vx,\pi) $.

Recall that \(\pi(i)\) is the decoding step of token \(x_i\). Thus,  \(x_{\pi^{-1}(j)}\) is the token decoded at step \(j\). Consider 
tokens 
decoded at steps $j$ and $j+1$, but 
are not in
descending total attention
order,
i.e.,
\begin{equation}\label{eq:order}
s_{\pi^{-1}(j)} < s_{\pi^{-1}(j+1)}.
\end{equation}
Consider the new permutation $\bar{\pi}$ with
$\bar{\pi}^{-1}(j+1)= \pi^{-1}(j),\; \bar{\pi}^{-1}(j)= \pi^{-1}(j+1)$,  and 
$\bar{\pi}^{-1}(k)= \pi^{-1}(k)$ for $k\neq j,j+1$.
Note that 
(\ref{eq:tmp})
can be written as
\begin{eqnarray*}
\widehat{\mathrm{UB}}(\vtheta,\vx,\pi)
& = &
\sqrt{2}\,\|\mW\|_F\,\frac{B}{b-1}\sum_{r=1}^b (r-1)\, s_{\pi^{-1}(r)} \\&= &
\sqrt{2}\,\|\mW\|_F\,\frac{B}{b-1}
\Bigl(
1\cdot s_{\pi^{-1}(2)}
+\cdots
+(j-1)\cdot s_{\pi^{-1}(j)}
+j\cdot s_{\pi^{-1}(j+1)}
+\cdots).
\end{eqnarray*}
Similarly,
\begin{align*}
\widehat{\mathrm{UB}}(\vtheta,\vx,\bar{\pi})
&=
\sqrt{2}\,\|\mW\|_F\,\frac{B}{b-1}
\Bigl(
1\cdot s_{\pi^{-1}(2)}
+\cdots
+j\cdot s_{\pi^{-1}(j)}
+(j-1)\cdot s_{\pi^{-1}(j+1)}
+\cdots
\Bigr).
\end{align*}
Therefore,
\begin{eqnarray*}
\lefteqn{\widehat{\mathrm{UB}}(\vtheta,\vx,\pi) -
\widehat{\mathrm{UB}}(\vtheta,\vx,\bar{\pi})}
\nonumber\\
&= &
\sqrt{2}\,\|\mW\|_F\,\frac{B}{b-1}
\Bigl(
(j-1)\, s_{\pi^{-1}(j)}
+
j\, s_{\pi^{-1}(j+1)}
-
j\, s_{\pi^{-1}(j)}
-
(j-1)\, s_{\pi^{-1}(j+1)}
\Bigr)
\nonumber\\
&=&
\sqrt{2}\,\|\mW\|_F\,\frac{B}{b-1}
\Bigl(
s_{\pi^{-1}(j+1)}-s_{\pi^{-1}(j)}
\Bigr).
\end{eqnarray*}
Thus, with  (\ref{eq:order}),
swapping the token pair order reduces $\widehat{\mathrm{UB}}(\vtheta,\vx,\pi)$.

Next, we prove the theorem by induction:

(i) Base Case: For a sequence of length $n=2$, the above shows that
the tokens should be decoded in descending 
total attention 
order.

(ii) Inductive Step: Assume that the hypothesis holds for a sequence of length $n-1$. We now consider a sequence of $n$ tokens. Let token $k$ be the element with the largest total attention.
We will show that token $k$ must be first decoded.
Assume, to the contrary, that token $k$ is decoded at step $q > 1$. By 
swapping it 
with the token decoded at step $q-1$, we can reduce $\widehat{\mathrm{UB}}(\vtheta,\vx,\pi)$ because token $k$ has the largest score. Therefore, in the optimal permutation $\pi$, token $k$ must be at position 1.
Once token $k$ is fixed at the first position, we are left with a sub-problem of the remaining $n-1$ tokens. By our inductive hypothesis, 
$\widehat{\mathrm{UB}}(\vtheta,\vx,\pi)$ 
for these $n-1$ tokens is minimized when they are arranged in descending total attention order.

\subsection{Proof of \cref{thm:conf_equivalent}}
\label{appendix_a4}
Let $\ell(\vz^\ast_i)$ be the confidence of token $i$ in the ideal prediction case, and $\ell(\vz_i)$ as the confidence in the real prediction case.
We first expand the  of a perturbed token using a first-order Taylor expansion:
\begin{equation} \label{eq:l}
\ell(\vz_i) = \ell(\vz_i^\ast) + \nabla \ell(\vz_i^\ast)^\top \vepsilon_i + O(|\vepsilon_i|^2),
\end{equation}
In the worst-case scenario, the error $\vepsilon_i$ is oriented in the direction of the gradient, i.e., $\vepsilon_i = -c \nabla \ell(\vz_i^\ast)$ for some $c > 0$. Substituting this into  (\ref{eq:l}), and dropping the 
$O(|\vepsilon_i|^2)$ term,
\begin{align*}
    \ell(\vz_i) 
    =
    \ell(\vz_i^\ast) - c \|\nabla \ell(\vz_i^\ast)\|_2 \cdot \|\vepsilon_i\|_2.
\end{align*}
By Assumption (iv), maximizing $\ell(\vz_i)$ is equivalent to minimizing the error magnitude $\|\vepsilon_i\|_2$:$$\arg\max_i \ell(\vz_i) = \arg\min_i \|\vepsilon_i\|_2.$$
By Assumption (i)
in \cref{thm:conf_equivalent},
$\|\vepsilon_i\|_2$ is a monotonically increasing function of the total
attention 
on masked  tokens.
Therefore:
$$\arg\min_i \|\vepsilon_i\|_2 = \arg\min_i \sum_{j \in \mathcal{M}} \mA_{ij}.$$
Since the total row sum of the attention matrix is normalized (i.e., $\sum_{j \in \mathcal{M}} \mA_{ij} + \sum_{j \notin \mathcal{M}} \mA_{ij} = 1$), 
$$\arg\min_i \sum_{j \in \mathcal{M}} \mA_{ij} = \arg\max_i \sum_{j \notin \mathcal{M}} \mA_{ij}.$$
Finally, applying assumption (ii), since $g$ is monotonically increasing, 
$$\arg\max_i \sum_{j \notin \mathcal{M}} \mA_{ij} = \arg\max_i g\left(\sum_{j \notin \mathcal{M}} \mA_{ij}\right) = \arg\max_i \sum_{j=1}^n \mA_{ji}.$$
This establishes the equivalence between the confidence-based selection and \dllma. 

\subsection{Proof of \cref{thm:conf_equivalent_entropy}}
\label{appendix_a5}
Let $\vp_i$ be the predictive distribution for token $i$, and define
\[
    p_i = \max_j [\vp_i]_j,
    \qquad
    n_i = -\mathcal H(\vp_i).
\]
Confidence-based sampling selects
\[
    i_{\mathrm{conf}}=\arg\max_i p_i,
\]
while entropy-based sampling selects
\[
    i_{\mathrm{ent}}=\arg\max_i n_i.
\]

\paragraph{Case (i).}
Suppose $p_{i^\ast}=1$ and $p_i<1$ for all $i\neq i^\ast$.
Since $p_{i^\ast}=1$, the distribution $\vp_{i^\ast}$ is deterministic.
Therefore,
\[
    \mathcal H(\vp_{i^\ast})=0,
    \qquad
    n_{i^\ast}=0.
\]
For any $i\neq i^\ast$, since $p_i<1$, the distribution $\vp_i$ is not
deterministic. Hence,
\[
    \mathcal H(\vp_i)>0,
    \qquad
    n_i=-\mathcal H(\vp_i)<0.
\]
Thus,
\[
    n_{i^\ast}>n_i,
    \qquad
    \forall i\neq i^\ast,
\]
which implies
\[
    i_{\mathrm{ent}}=i^\ast.
\]
Moreover, since $p_{i^\ast}=1$ and $p_i<1$ for all $i\neq i^\ast$, we also have
\[
    i_{\mathrm{conf}}=i^\ast.
\]
Therefore,
\[
    i_{\mathrm{ent}}=i_{\mathrm{conf}}=i^\ast.
\]

\paragraph{Case (ii).}
Suppose $p_{i^\ast}>p_i$ for all $i\neq i^\ast$, and
\[
    -\mathcal H(\vp_{i^\ast})>\log p_i,
    \qquad
    \forall i\neq i^\ast.
\]
We first show that, for any token $i$,
\[
    -\mathcal H(\vp_i)\le \log p_i.
\]
Since $p_i=\max_j[\vp_i]_j$, we have
\[
    [\vp_i]_j\le p_i,
    \qquad
    \forall j.
\]
Therefore,
\[
    \log \frac{1}{[\vp_i]_j}
    \ge
    \log \frac{1}{p_i},
    \qquad
    \forall j.
\]
Using the definition of entropy,
\[
\begin{aligned}
    \mathcal H(\vp_i)
    &=
    \sum_j [\vp_i]_j
    \log \frac{1}{[\vp_i]_j} \\
    &\ge
    \sum_j [\vp_i]_j
    \log \frac{1}{p_i} \\
    &=
    \log \frac{1}{p_i}.
\end{aligned}
\]
Thus,
\[
    -\mathcal H(\vp_i)
    \le
    \log p_i.
\]
That is,
\[
    n_i\le \log p_i.
\]

For any $i\neq i^\ast$, the assumption gives
\[
    n_{i^\ast}
    =
    -\mathcal H(\vp_{i^\ast})
    >
    \log p_i.
\]
Combining this with $n_i\le \log p_i$, we obtain
\[
    n_{i^\ast}>n_i,
    \qquad
    \forall i\neq i^\ast.
\]
Hence,
\[
    i_{\mathrm{ent}}=\arg\max_i n_i=i^\ast.
\]
Meanwhile, since $p_{i^\ast}>p_i$ for all $i\neq i^\ast$, confidence-based
sampling also selects
\[
    i_{\mathrm{conf}}=\arg\max_i p_i=i^\ast.
\]
Therefore,
\[
    i_{\mathrm{ent}}=i_{\mathrm{conf}}=i^\ast.
\]

This completes the proof.

\section{Broader impacts}
This work improves the inference efficiency of dLLMs, which may enhance the accessibility and energy efficiency of large language models by reducing computational cost. Our work is primarily methodological and does not introduce new application-specific deployment scenarios. While it may influence downstream research and applications, we do not identify any direct negative societal impacts beyond the general risks associated with broader deployment of language models.

\section{Limitation}
This work studies the sampling order selection problem in Transformer-based masked diffusion language models and introduce the Attn-sampler algorithm.
One of the limitation is that current Attn-sampler cannot be directly extended to uniform diffusion.
In uniform diffusion, the decoding process is more complex, since each token may be updated multiple times throughout sampling. 
Our Attn-Sampler is based on an optimal permutation, which relies on the one-pass decoding structure of masked dLLMs.
Extending the framework to uniform diffusion is an interesting direction, and we will leave it for future work.

\section*{NeurIPS Paper Checklist}

\begin{enumerate}

\item {\bf Claims}
    \item[] Question: Do the main claims made in the abstract and introduction accurately reflect the paper's contributions and scope?
    \item[] Answer: \answerYes{} 
    \item[] Justification: Main claims made in the abstract and introduction accurately reflect the paper's contributions and scope.

\item {\bf Limitations}
    \item[] Question: Does the paper discuss the limitations of the work performed by the authors?
    \item[] Answer: \answerYes{} 
    \item[] Justification: Limitations are discussed in Appendix C of the main paper.

\item {\bf Theory assumptions and proofs}
    \item[] Question: For each theoretical result, does the paper provide the full set of assumptions and a complete (and correct) proof?
    \item[] Answer: \answerYes{} 
    \item[] Justification: All assumptions and proofs are provided.

    \item {\bf Experimental result reproducibility}
    \item[] Question: Does the paper fully disclose all the information needed to reproduce the main experimental results of the paper to the extent that it affects the main claims and/or conclusions of the paper (regardless of whether the code and data are provided or not)?
    \item[] Answer: \answerYes{} 
    \item[] Justification: Algorithms~1 and~2 in the main paper describe the proposed method in detail. Details of setup are provided in the experiment section.

\item {\bf Open access to data and code}
    \item[] Question: Does the paper provide open access to the data and code, with sufficient instructions to faithfully reproduce the main experimental results, as described in supplemental material?
    \item[] Answer: \answerNo{} 
    \item[] Justification: The code will be made publicly available after the paper is accepted.

\item {\bf Experimental setting/details}
    \item[] Question: Does the paper specify all the training and test details (e.g., data splits, hyperparameters, how they were chosen, type of optimizer) necessary to understand the results?
    \item[] Answer: \answerYes{} 
    \item[] Justification: Discussed in Setup section of the main paper.

\item {\bf Experiment statistical significance}
    \item[] Question: Does the paper report error bars suitably and correctly defined or other appropriate information about the statistical significance of the experiments?
    \item[] Answer: \answerNo{} 
    \item[] Justification: Our experiments are deterministic under the fixed evaluation protocol.

\item {\bf Experiments compute resources}
    \item[] Question: For each experiment, does the paper provide sufficient information on the computer resources (type of compute workers, memory, time of execution) needed to reproduce the experiments?
    \item[] Answer: \answerYes{} 
    \item[] Justification: Experiments compute resources are included in Experiment section of the main paper.
    
\item {\bf Code of ethics}
    \item[] Question: Does the research conducted in the paper conform, in every respect, with the NeurIPS Code of Ethics \url{https://neurips.cc/public/EthicsGuidelines}?
    \item[] Answer: \answerYes{} 
    \item[] Justification: The research presented in this paper conforms to the NeurIPS Code of Ethics.

\item {\bf Broader impacts}
    \item[] Question: Does the paper discuss both potential positive societal impacts and negative societal impacts of the work performed?
    \item[] Answer: \answerYes{}
    \item[] Justification: Broader impacts are included in appendix.
    
\item {\bf Safeguards}
    \item[] Question: Does the paper describe safeguards that have been put in place for responsible release of data or models that have a high risk for misuse (e.g., pre-trained language models, image generators, or scraped datasets)?
    \item[] Answer: \answerNA{}
    \item[] Justification: The paper poses no such risks. 

\item {\bf Licenses for existing assets}
    \item[] Question: Are the creators or original owners of assets (e.g., code, data, models), used in the paper, properly credited and are the license and terms of use explicitly mentioned and properly respected?
    \item[] Answer: \answerYes{} 
    \item[] Justification: We use publicly available models and datasets, and cite their original creators or owners throughout the paper. Their licenses and terms of use are reviewed and respected.

\item {\bf New assets}
    \item[] Question: Are new assets introduced in the paper well documented and is the documentation provided alongside the assets?
    \item[] Answer: \answerNA{} 
    \item[] Justification: The paper does not release new assets.

\item {\bf Crowdsourcing and research with human subjects}
    \item[] Question: For crowdsourcing experiments and research with human subjects, does the paper include the full text of instructions given to participants and screenshots, if applicable, as well as details about compensation (if any)? 
    \item[] Answer: \answerNA{} 
    \item[] Justification: The paper does not involve crowdsourcing nor research with human subjects.

\item {\bf Institutional review board (IRB) approvals or equivalent for research with human subjects}
    \item[] Question: Does the paper describe potential risks incurred by study participants, whether such risks were disclosed to the subjects, and whether Institutional Review Board (IRB) approvals (or an equivalent approval/review based on the requirements of your country or institution) were obtained?
    \item[] Answer: \answerNA{} 
    \item[] Justification: The paper does not involve crowdsourcing nor research with human subjects.

\item {\bf Declaration of LLM usage}
    \item[] Question: Does the paper describe the usage of LLMs if it is an important, original, or non-standard component of the core methods in this research? Note that if the LLM is used only for writing, editing, or formatting purposes and does \emph{not} impact the core methodology, scientific rigor, or originality of the research, declaration is not required.
    \item[] Answer: \answerNA{} 
    \item[] Justification: The core method development in this research does not involve LLMs as any important, original, or non-standard components.

\end{enumerate}

\end{document}